\documentclass[11pt,letterpaper]{article}
\usepackage{emnlp2017}
\usepackage{times}
\usepackage{latexsym}

\usepackage{graphicx}
\usepackage{amssymb}
\usepackage{amsthm}
\usepackage{mathtools}
\usepackage{enumerate}
\usepackage{lettrine}

\emnlpfinalcopy

\def\emnlppaperid{59}

\usepackage[utf8]{inputenc}
\usepackage[T1]{fontenc}

\usepackage{multirow}

\usepackage{tikz}
\usetikzlibrary{arrows, backgrounds, positioning, fit, petri,
  patterns, shadows, calc, intersections,
  shapes.multipart}

\tikzstyle{ublock} = [rectangle, text centered, rounded corners, minimum height=2em, minimum width=2.5em]
\tikzstyle{block} = [draw, ublock]
\tikzstyle{double} = [minimum width=4em]

\definecolor{left}{HTML}{336B87}
\definecolor{right}{HTML}{763626}

\tikzstyle{fwblock} = [draw, ublock, fill=left!70]
\tikzstyle{bwblock} = [draw, ublock, fill=right]
\tikzstyle{mergedblock} = [block, rectangle split, rectangle split horizontal, rectangle split parts=2, rectangle split part fill={left!70,right}, rectangle split draw splits=false]

\usepackage{pgfplots}
\usepackage{amssymb}

\usepackage{color}

\title{Deep Architectures for Neural Machine Translation}

\author{
Antonio Valerio Miceli Barone$^\dag$ \quad Jindřich Helcl$^\star$ \quad Rico Sennrich$^\dag$\\
\bf Barry Haddow$^\dag$ \quad Alexandra Birch$^\dag$\\
$^\dag$School of Informatics, University of Edinburgh\\
$^\star$Faculty of Mathematics and Physics, Charles University\\
{\tt \{amiceli, bhaddow\}@inf.ed.ac.uk}\\
{\tt \{rico.sennrich, a.birch\}@ed.ac.uk}\\
{\tt helcl@ufal.mff.cuni.cz}\\
}

\date{}

\begin{document}
\maketitle
\begin{abstract}

It has been shown that increasing model depth improves the quality of neural machine translation.
However, different architectural variants to increase model depth have been proposed, and so far, there has been no thorough comparative study.

In this work, we describe and evaluate several existing approaches to introduce depth in neural machine translation.
Additionally, we explore novel architectural variants, including deep transition RNNs, and we vary how attention is used in the deep decoder.
We introduce a novel "BiDeep" RNN architecture that combines deep transition RNNs and stacked RNNs.

Our evaluation is carried out on the English to German WMT news translation dataset, using a single-GPU machine for both training and inference.
We find that several of our proposed architectures improve upon existing approaches in terms of speed and translation quality.
We obtain best improvements with a BiDeep RNN of combined depth 8, obtaining an average improvement of 1.5 {\sc Bleu} over a strong shallow baseline.

We release our code for ease of adoption.

\end{abstract}

\section{Introduction}
\label{sec:intro}

Neural machine translation (NMT) is a well-established approach that yields the
best results on most language pairs \citep{bojar-EtAl:2016:WMT1,iwslt-report}.
Most systems are based on the sequence-to-sequence model with attention \citep{bahdanau2014neural} which employs single-layer recurrent neural networks both in the encoder and in the decoder.

Unlike feed-forward networks where depth is straightforwardly defined as the number of non-input layers, recurrent neural network architectures with multiple layers allow different connection schemes \citep{Pascanu+et+al-ICLR2014} that give rise to different, orthogonal, definitions of depth \citep{DBLP:journals/corr/ZhangWCLMSB16} which can affect the model performance depending on a given task.
This is further complicated in sequence-to-sequence models as they contain multiple sub-networks, recurrent or feed-forward, each of which can be deep in different ways, giving rise to a large number of possible configurations.

In this work we focus on \textit{stacked} and \textit{deep transition} recurrent architectures as defined by \citet{Pascanu+et+al-ICLR2014}.
Different types of stacked architectures have been successfully used for NMT  \citep{zhou2016deep, wu2016google}.
However, there is a lack of empirical comparisons of different deep architectures.
Deep transition architectures have been successfully used for language modeling \citep{zilly2016recurrent}, but not for NMT so far.
We evaluate these architectures, both alone and in combination, varying the connection scheme between the different components and their depth over the different dimensions, measuring the  performance of the different configurations on the WMT news translation task.\footnote{\url{http://www.statmt.org/wmt17/translation-task.html}}

Related work includes that of \citet{DBLP:journals/corr/BritzGLL17}, who have performed an exploration of NMT architectures in parallel to our work.
Their experiments, which are largely orthogonal to ours, focus on embedding size, RNN cell type (GRU vs. LSTM), network depth (defined according to the architecture of \citet{wu2016google}), attention mechanism and beam size.
\citet{DBLP:journals/corr/GehringAGYD17} recently proposed a NMT architecture based on convolutions over fixed-sized windows rather than RNNs, and they reported results for different model depths and attention mechanism configurations.
A similar feedforward architecture which uses multiple pervasive attention mechanisms rather than convolutions was proposed by \citet{vaswani2017attention}, who also report results for different model depths.

\section{NMT Architectures}
\label{sec:deepnmt}

All the architectures that we consider in this work are GRU \citep{DBLP:journals/corr/ChoMBB14}  sequence-to-sequence transducers \citep{sutskever2014sequence, cho2014learning} with attention \citep{bahdanau2014neural}.
In this section we describe the baseline system and the variants that we evaluated.

\subsection{Baseline Architecture}

As our baseline, we use the NMT architecture implemented in Nematus, which is described in more depth by \citet{sennrich-EtAl:2017:EACLDemo}.
We augment it with layer normalization \citep{DBLP:journals/corr/BaKH16}, which we have found to both improve translation quality and make training considerably faster.

For our discussion, it is relevant that the baseline architecture already exhibits two types of depth:

\begin{itemize}
\item \textit{recurrence transition depth} in the decoder RNN which consists of two GRU transitions per output word with an attention mechanism in between, as described in \citet{firat2016cgru}.
\item \textit{feed-forward depth} in the attention network that computes the alignment scores and in the output network that predicts the target words.
Both these networks are multi-layer perceptrons with one $\text{tanh}$ hidden layer.
\end{itemize}

\subsection{Deep Transition Architectures}
\label{drt-section}

In a deep transition RNN (DT-RNN), at each time step the next state is computed by the sequential application of multiple transition layers, effectively using a feed-forward network embedded inside the recurrent cell.
In our experiments, these layers are GRU transition blocks with independently trainable parameters, connected such that the "state" output of one of them is used as the "state" input of the next one.
Note that each of these GRU transition is not individually recurrent, recurrence only occurs at the level of the whole multi-layer cell, as the "state" output of the last GRU transition for the current time step is carried over as the "state" input of the first GRU transition for the next time step.

Applying this architecture to NMT is a novel contribution.

\subsubsection{Deep Transition Encoder}
\label{drt-enc-section}

As in a baseline shallow Nematus system, the encoder is a bidirectional recurrent neural network.
Let $L_s$ be the encoder recurrence depth, then for the $i$-th source word in the forward direction the forward source word state $\overrightarrow{h}_i \equiv \overrightarrow{h}_{i,L_s}$ is computed as:
\begin{align*}
\overrightarrow{h}_{i,1} &= \text{GRU}_1   \left( x_{i}, \overrightarrow{h}_{i-1,L_s}  \right) \\
\overrightarrow{h}_{i,k} &= \text{GRU}_k   \left(0, \overrightarrow{h}_{i,k-1}  \right) \text{for } 1 < k \leq L_s
\end{align*}
where the input to the first GRU transition is the word embedding $x_{i}$, while the other GRU transitions have no external inputs.
Recurrence occurs as the previous word state $\overrightarrow{h}_{i-1,L_s}$ enters the computation in the first GRU transition for the current word. \\
The reverse source word states are computed similarly and concatenated to the forward ones to form the bidirectional source word states $C \equiv \left\lbrace \left[\overrightarrow{h}_{i,L_s} \overleftarrow{h}_{i,L_s} \right] \right\rbrace$.

\subsubsection{Deep Transition Decoder}
\label{drt-dec-section}

\begin{figure}
\centering
\begin{tikzpicture}[scale=0.8, transform shape]

  \node[fwblock](d11){};
  \node[fwblock, right=1em of d11](d12){};
  \node[ublock, right=1em of d12](d1d){$\ldots$};
  \node[fwblock, right=1em of d1d](d1n){};


  \node[fwblock, above=1em of d11](d21){};
  \node[fwblock, right=1em of d21](d22){};
  \node[ublock, right=1em of d22](d2d){$\ldots$};
  \node[fwblock, right=1em of d2d](d2n){};


  \draw[-latex', thick] (d11.north) -- (d21.south);
  \draw[-latex', thick] (d12.north) -- (d22.south);
  \draw[-latex', thick] (d1n.north) -- (d2n.south);

  \node[fwblock, above=1em of d21](d31){};
  \node[fwblock, right=1em of d31](d32){};
  \node[ublock, right=1em of d32](d3d){$\ldots$};
  \node[fwblock, right=1em of d3d](d3n){};


  \draw[-latex', thick] (d21.north) -- (d31.south);
  \draw[-latex', thick] (d22.north) -- (d32.south);
  \draw[-latex', thick] (d2n.north) -- (d3n.south);

  \draw[-latex', thick] (d31.north) to[out=60,in=-120, looseness=1.6] (d12.south);
  \draw[-latex', thick] (d32.north) to[out=60,in=-120, looseness=1.6] (d1d.south);
  \draw[-latex', thick] (d3d.north) to[out=60,in=-120, looseness=1.6] (d1n.south);
%
%
%
%
%
%

\end{tikzpicture}
\caption{Deep transition decoder}
\label{drt-dec}
\end{figure}
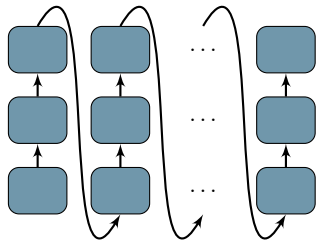

The deep transition decoder is obtained by extending the baseline decoder in a similar way.
Recall that the baseline decoder of Nematus already has a transition depth of two, with the first GRU transition receiving as input the embedding of the previous target word and the second GRU transition receiving as input a context vector computed by the attention mechanism.
We extend this decoder architecture to an arbitrary transition depth $L_t$ as follows:
\begin{align*}
s_{j,1} &= \text{GRU}_1   \left( y_{j-1}, s_{j-1,L_t}  \right) \\
s_{j,2} &= \text{GRU}_2   \left( \text{ATT}(C, s_{j,1}), s_{j,1}  \right) \\
s_{j,k} &= \text{GRU}_k   \left(0, s_{j,k-1}  \right) \text{for } 2 < k \leq L_t
\end{align*}
where $y_{j-1}$ is the embedding of the previous target word and $\text{ATT}(C, s_{i,1})$ is the context vector computed by the attention mechanism.
GRU transitions other than the first two do not have external inputs.
The target word state vector $s_j \equiv s_{j, L_t}$ is then used by the feed-forward output network to predict the current target word.
A diagram of this architecture is shown in Figure~\ref{drt-dec}.

The output network can be also made deeper by adding more feed-forward hidden layers.

\subsection{Stacked architectures}

A stacked RNN is obtained by having multiple RNNs (GRUs in our experiments) run for the same number of time steps, connected such that at each step the bottom RNN takes "external" inputs from the outside, while each of the higher RNN takes as its "external" input the "state" output of the one below it.
Residual connections between states at different depth \citep{DBLP:conf/cvpr/HeZRS16} are also used to improve information flow.
Note that unlike deep transition GRUs, here each GRU transition block constitutes a cell that is individually recurrent, as it has its own state that is carried over between time steps.

\subsubsection{Stacked Encoder}

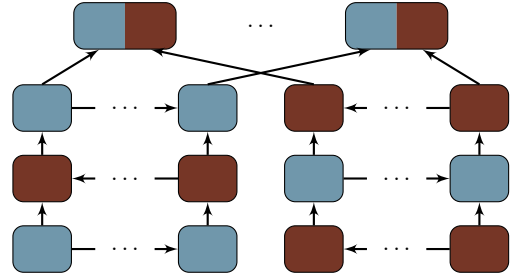
\begin{figure}
\centering
  \begin{tikzpicture}[scale=0.8, transform shape]

  \node[fwblock](f11){};
  \node[ublock, right=1 em of f11](f1d){$\ldots$};
  \node[fwblock, right=1em of f1d](f1n){};

  \draw[thick] (f11.east) -- (f1d.west);
  \draw[-latex', thick] (f1d.east) -- (f1n.west);

  \node[bwblock, right=2em of f1n](r11){};
  \node[ublock, right=1em of r11](r1d){$\ldots$};
  \node[bwblock, right=1em of r1d](r1n){};

  \draw[-latex', thick] (r1d.west) -- (r11.east);
  \draw[thick] (r1n.west) -- (r1d.east);

  \node[bwblock, above=1em of f11](r21){};
  \node[ublock, right=1em of r21](r2d){$\ldots$};
  \node[bwblock, right=1em of r2d](r2n){};

  \draw[-latex', thick] (r2d.west) -- (r21.east);
  \draw[thick] (r2n.west) -- (r2d.east);

  \node[fwblock, right=2em of r2n](f21){};
  \node[ublock, right=1 em of f21](f2d){$\ldots$};
  \node[fwblock, right=1em of f2d](f2n){};

  \draw[thick] (f21.east) -- (f2d.west);
  \draw[-latex', thick] (f2d.east) -- (f2n.west);

  \draw[-latex', thick] (f11.north) -- (r21.south);
  \draw[-latex', thick] (f1n.north) -- (r2n.south);

  \draw[-latex', thick] (r11.north) -- (f21.south);
  \draw[-latex', thick] (r1n.north) -- (f2n.south);

  \node[fwblock, above=1em of r21](f31){};
  \node[ublock, right=1em of f31](f3d){$\ldots$};
  \node[fwblock, right=1em of f3d](f3n){};

  \draw[thick] (f31.east) -- (f3d.west);
  \draw[-latex', thick] (f3d.east) -- (f3n.west);

  \node[bwblock, right=2em of f3n](r31){};
  \node[ublock, right=1 em of r31](r3d){$\ldots$};
  \node[bwblock, right=1em of r3d](r3n){};

  \draw[-latex', thick] (r3d.west) -- (r31.east);
  \draw[thick] (r3n.west) -- (r3d.east);

  \draw[-latex', thick] (r21.north) -- (f31.south);
  \draw[-latex', thick] (r2n.north) -- (f3n.south);

  \draw[-latex', thick] (f21.north) -- (r31.south);
  \draw[-latex', thick] (f2n.north) -- (r3n.south);

  \node[mergedblock, double, above=1.5em of f3d](c11){\nodepart[text width=1.5em]{one}\nodepart[text width=1.5em]{two}};
  \node[mergedblock, double, above=1.5em of r3d](c1n){\nodepart[text width=1.5em]{one}\nodepart[text width=1.5em]{two}};
  \node[ublock](c1d) at ($(c11)!.5!(c1n)$) {$\ldots$};


  \draw[-latex', thick] (f31.north) -- (c11.one south);
  \draw[-latex', thick] (r31.north) -- (c11.two south);

  \draw[-latex', thick] (f3n.north) -- (c1n.one south);
  \draw[-latex', thick] (r3n.north) -- (c1n.two south);

\end{tikzpicture}
\caption{Alternating stacked encoder \citep{zhou2016deep}.}
\label{baidu-enc}
\end{figure}

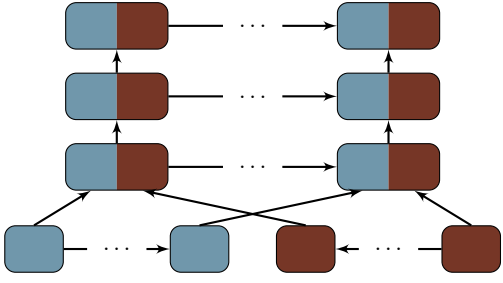
\begin{figure}
\centering
\begin{tikzpicture}[scale=0.8, transform shape]

  \node[fwblock](f11){};
  \node[ublock, right=1 em of f11](f1d){$\ldots$};
  \node[fwblock, right=1em of f1d](f1n){};

  \draw[thick] (f11.east) -- (f1d.west);
  \draw[-latex', thick] (f1d.east) -- (f1n.west);

  \node[bwblock, right=2em of f1n](r11){};
  \node[ublock, right=1em of r11](r1d){$\ldots$};
  \node[bwblock, right=1em of r1d](r1n){};

  \draw[-latex', thick] (r1d.west) -- (r11.east);
  \draw[thick] (r1n.west) -- (r1d.east);

  \node[mergedblock, double, above=1.5em of f1d](c11){\nodepart[text width=1.5em]{one}\nodepart[text width=1.5em]{two}};
  \node[mergedblock, double, above=1.5em of r1d](c1n){\nodepart[text width=1.5em]{one}\nodepart[text width=1.5em]{two}};
  \node[ublock](c1d) at ($(c11)!.5!(c1n)$) {$\ldots$};

  \draw[thick] (c11.east) -- (c1d.west);
  \draw[-latex', thick] (c1d.east) -- (c1n.west);

  \node[mergedblock, double, above=1em of c11](c21){\nodepart[text width=1.5em]{one}\nodepart[text width=1.5em]{two}};
  \node[mergedblock, double, above=1em of c1n](c2n){\nodepart[text width=1.5em]{one}\nodepart[text width=1.5em]{two}};
  \node[ublock](c2d) at ($(c21)!.5!(c2n)$) {$\ldots$};

  \draw[thick] (c21.east) -- (c2d.west);
  \draw[-latex', thick] (c2d.east) -- (c2n.west);

  \draw[-latex', thick] (c11.north) -- (c21.south);
  \draw[-latex', thick] (c1n.north) -- (c2n.south);

  \node[mergedblock, double, above=1em of c21](c31){\nodepart[text width=1.5em]{one}\nodepart[text width=1.5em]{two}};
  \node[mergedblock, double, above=1em of c2n](c3n){\nodepart[text width=1.5em]{one}\nodepart[text width=1.5em]{two}};
  \node[ublock](c3d) at ($(c31)!.5!(c3n)$) {$\ldots$};

  \draw[thick] (c31.east) -- (c3d.west);
  \draw[-latex', thick] (c3d.east) -- (c3n.west);

  \draw[-latex', thick] (c21.north) -- (c31.south);
  \draw[-latex', thick] (c2n.north) -- (c3n.south);

  \draw[-latex', thick] (f11.north) -- (c11.one south);
  \draw[-latex', thick] (r11.north) -- (c11.two south);

  \draw[-latex', thick] (f1n.north) -- (c1n.one south);
  \draw[-latex', thick] (r1n.north) -- (c1n.two south);

\end{tikzpicture}
\caption{Biunidirectional stacked encoder \citep{wu2016google}.}
\label{google-enc}
\end{figure}

In this work we consider two types of bidirectional stacked encoders: an architecture similar to \citet{zhou2016deep} which we denote here as \textit{alternating} encoder (Figure~\ref{baidu-enc}), and one similar to \citet{wu2016google} which we denote as \textit{biunidirectional} encoder (Figure~\ref{google-enc}).

Our contribution is the empirical comparison of these architectures, both in isolation and in combination with the deep transition architecture.

We do not consider stacked unidirectional encoders \citep{sutskever2014sequence} as bidirectional encoders have been shown to outperform them (e.g. \citet{DBLP:journals/corr/BritzGLL17}).

\paragraph{Alternating Stacked Encoder}

The forward part of the encoder consists of a stack of GRU recurrent neural networks, the first one processing words in the forward direction, the second one in the backward direction, and so on, in alternating directions.
For an encoder stack depth $D_s$, and a source sentence length $N$, the forward source word state $\overrightarrow{w}_i \equiv \overrightarrow{w}_{i,D_s}$ is computed as:
\begin{align*}
\overrightarrow{w}_{i,1} &= \overrightarrow{h}_{i,1} = \text{GRU}_1   \left( x_{i}, \overrightarrow{h}_{i-1,1}  \right) \\
\overrightarrow{h}_{i,2k} &= \text{GRU}_{2k}   \left(\overrightarrow{w}_{i,2k-1}, \overrightarrow{h}_{i+1,2k}  \right) \\
  &\text{ for } 1 < 2k \leq D_s \\
\overrightarrow{h}_{i,2k+1} &= \text{GRU}_{2k+1}   \left(\overrightarrow{w}_{i,2k}, \overrightarrow{h}_{i-1,2k+1}  \right) \\
  &\text{ for } 1 < 2k+1 \leq D_s \\
\overrightarrow{w}_{i,j} &= \overrightarrow{h}_{i,j} + \overrightarrow{w}_{i,j-1} \\
& \text{ for } 1 < j \leq D_s
\end{align*}
where we assume that $\overrightarrow{h}_{0,k}$ and $\overrightarrow{h}_{N+1,k}$ are zero vectors.
Note the residual connections: at each level above the first one, the word state of the previous level $\overrightarrow{w}_{i,j-1}$ is added to the recurrent state of the GRU cell $\overrightarrow{h}_{i,j}$ to compute the the word state for the current level $\overrightarrow{w}_{i,j}$.

The backward part of the encoder has the same structure, except that the first level of the stack processes the words in the backward direction and the subsequent levels alternate directions.

The forward and backward word states are then concatenated to form bidirectional word states $C \equiv \left\lbrace \left[\overrightarrow{w}_{i,D_s} \overleftarrow{w}_{i,D_s} \right] \right\rbrace$.
A diagram of this architecture is shown in Figure~\ref{baidu-enc}.

\paragraph{Biunidirectional Stacked Encoder}

In this encoder the forward and backward parts are shallow, as in the baseline architecture.
Their word states are concatenated to form shallow bidirectional word states $ w_{i} \equiv \left[\overrightarrow{w}_{i,1} \overleftarrow{w}_{i,1}\right]$ that are then used as inputs for subsequent stacked GRUs which operate only in the forward sentence direction, hence the name "biunidirectional".
Since residual connections are also present, the higher depth GRUs have a state size twice that of the base ones.
This architecture has shorter maximum information propagation paths than the alternating encoder, suggesting that it may be less expressive, but it has the advantage of enabling implementations with higher model parallelism.
A diagram of this architecture is shown in Figure~\ref{google-enc}.

In principle, alternating and biunidirectional stacked encoders can be combined by having $D_{sa}$ alternating layers followed by $D_{sb}$ unidirectional layers.

\subsubsection{Stacked Decoder}

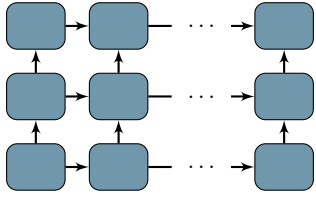
\begin{figure}
\centering
\begin{tikzpicture}[scale=0.8, transform shape]

  \node[fwblock](d11){};
  \node[fwblock, right=1em of d11](d12){};
  \node[ublock, right=1em of d12](d1d){$\ldots$};
  \node[fwblock, right=1em of d1d](d1n){};

  \draw[-latex', thick] (d11.east) -- (d12.west);
  \draw[thick] (d12.east) -- (d1d.west);
  \draw[-latex', thick] (d1d.east) -- (d1n.west);

  \node[fwblock, above=1em of d11](d21){};
  \node[fwblock, right=1em of d21](d22){};
  \node[ublock, right=1em of d22](d2d){$\ldots$};
  \node[fwblock, right=1em of d2d](d2n){};

  \draw[-latex', thick] (d21.east) -- (d22.west);
  \draw[thick] (d22.east) -- (d2d.west);
  \draw[-latex', thick] (d2d.east) -- (d2n.west);

  \draw[-latex', thick] (d11.north) -- (d21.south);
  \draw[-latex', thick] (d12.north) -- (d22.south);
  \draw[-latex', thick] (d1n.north) -- (d2n.south);

  \node[fwblock, above=1em of d21](d31){};
  \node[fwblock, right=1em of d31](d32){};
  \node[ublock, right=1em of d32](d3d){$\ldots$};
  \node[fwblock, right=1em of d3d](d3n){};

  \draw[-latex', thick] (d31.east) -- (d32.west);
  \draw[thick] (d32.east) -- (d3d.west);
  \draw[-latex', thick] (d3d.east) -- (d3n.west);

  \draw[-latex', thick] (d21.north) -- (d31.south);
  \draw[-latex', thick] (d22.north) -- (d32.south);
  \draw[-latex', thick] (d2n.north) -- (d3n.south);

%
%
%
%
%
%

\end{tikzpicture}
\caption{Stacked RNN decoder}
\label{s-dec}
\end{figure}

A stacked decoder can be obtained by stacking RNNs which operate in the forward sentence direction.
A diagram of this architecture is shown in Figure~\ref{s-dec}.

Note that the base RNN is always a conditional GRU (cGRU, \citeauthor{firat2016cgru}, \citeyear{firat2016cgru}) which has transition depth at least two due to the way that the context vectors generated by the attention mechanism are used in Nematus.
This opens up the possibility of several architectural variants which we evaluated in this work:

\paragraph{Stacked GRU}
The higher RNNs are simple GRUs which receive as input the state from the previous level of the stack, with residual connections between the levels.
\begin{align*}
s_{j,1,1} &= \text{GRU}_{1,1}   \left( y_{j-1}, s_{j-1,1,2}  \right) \\
c_{j,1} &= \text{ATT}(C, s_{j,1,1}) \\
s_{j,1,2} &= \text{GRU}_{1,2} \left(c_{j,1}, s_{j,1,1}  \right) \\
r_{j,1} &= s_{j,1,2} \\
s_{j,k,1} &= \text{GRU}_k   \left(r_{j,k-1}, s_{j-1,k,1}  \right) \\
r_{j,k} &= s_{j,k,1} + r_{j,k-1} \\
& \text{for } 1 < k \leq D_t \\
\end{align*}
Note that the higher levels have transition depth one, unlike the base level which has two.

\paragraph{Stacked rGRU}
The higher RNNs are GRUs whose "external" input is the concatenation of the state below and the context vector from the base RNN.
Formally, the states $s_{j,k,1}$ of the higher RNNs are computed as:
\begin{align*}
s_{j,k,1} &= \text{GRU}_k   \left(\left[r_{j,k-1}, c_{j,1} \right], s_{j-1,k,1}  \right) \\
& \text{for } 1 < k \leq D_t \\
\end{align*}
This is similar to the deep decoder by \citet{wu2016google}.

\paragraph{Stacked cGRU}
The higher RNNs are conditional GRUs, each with an independent attention mechanism.
Each level has two GRU transitions per step $j$, with a new context vector $c_{j,k}$ computed in between:
\begin{align*}
s_{j,k,1} &= \text{GRU}_{k,1}   \left(r_{j,k-1}, s_{j-1,k,1}  \right) \\
c_{j,k} &= \text{ATT}(C, s_{j,k,1}) \\
s_{j,k,2} &= \text{GRU}_{k,2} \left(c_{j,k}, s_{j,1,1}  \right) \\
& \text{for } 1 < k \leq D_t \\
\end{align*}
Note that unlike the stacked GRU and rGRU, the higher levels have transition depth two.

\paragraph{Stacked crGRU}
The higher RNNs are conditional GRUs but they reuse the context vectors from the base RNN.
Like the cGRU there are two GRU transition per step, but they reuse the context vector $c_{j,1}$ computed at the first level of the stack:
\begin{align*}
s_{j,k,1} &= \text{GRU}_{k,1} \left(r_{j,k-1}, s_{j-1,k,1}  \right) \\
s_{j,k,2} &= \text{GRU}_{k,2} \left(c_{j,1}, s_{j,1,1}  \right) \\
& \text{for } 1 < k \leq D_t \\
\end{align*}

\subsection{BiDeep architectures}

We introduce the \textit{BiDeep RNN}, a novel architecture obtained by combining deep transitions with stacking.

A BiDeep encoder is obtained by replacing the $D_s$ individually recurrent GRU cells of a stacked encoder with multi-layer deep transition cells each composed by $L_s$ GRU transition blocks.

For instance, the BiDeep alternating encoder is defined as follows:
\begin{align*}
\overrightarrow{w}_{i,1} &= \overrightarrow{h}_{i,1} = \text{DTGRU}_1   \left( x_{i}, \overrightarrow{h}_{i-1,1}  \right) \\
\overrightarrow{h}_{i,2k} &= \text{DTGRU}_{2k}   \left(\overrightarrow{w}_{i,2k-1}, \overrightarrow{h}_{i+1,2k}  \right) \\
  &\text{ for } 1 < 2k \leq D_s \\
\overrightarrow{h}_{i,2k+1} &= \text{DTGRU}_{2k+1}   \left(\overrightarrow{w}_{i,2k}, \overrightarrow{h}_{i-1,2k+1}  \right) \\
  &\text{ for } 1 < 2k+1 \leq D_s \\
\overrightarrow{w}_{i,j} &= \overrightarrow{h}_{i,j} + \overrightarrow{w}_{i,j-1} \\
& \text{ for } 1 < j \leq D_s
\end{align*}
where each multi-layer cell $\text{DTGRU}_k$ is defined as:
\begin{align*}
v_{k,1} &= \text{GRU}_{k, 1} \left(\text{in}_k, \text{state}_k  \right) \\
v_{k,t} &= \text{GRU}_{k, t} \left(0, v_{k_t-1} \right)
\text{for } 1 < k \leq L_s \\
\text{DTGRU}_k &\left(\text{in}_k, \text{state}_k \right) = v_{k,L_s}
\end{align*}

It is also possible to have different transition depths at each stacking level.

BiDeep decoders are similarly defined, replacing the recurrent cells (GRU, rGRU, cGRU or crGRU) with deep transition multi-layer cells.

\section{Experiments}
\label{sec:experiments}

All experiments were performed with Nematus \citep{sennrich-EtAl:2017:EACLDemo},
following \citet{uedin-nmt:2017} in their choice of preprocessing and hyperparameters.
For experiments with deep models, we increase the depth by a factor of 4 compared to the baseline for most experiments;
in preliminary experiments, we observed diminishing returns for deeper models.

We trained on the parallel English--German training data of WMT-2017 news translation task, using newstest2013 as validation set.
We used early-stopping on the validation cross-entropy and selected the best model based on validation {\sc Bleu}.

We report cross-entropy (CE) on newstest2013, training speed (on a single Titan X (Pascal) GPU), and the number of parameters.
For translation quality, we report case-sensitive, detokenized {\sc Bleu}, measured with mteval-v13a.pl, on newstest2014, newstest2015, and newstest2016.

We release the code under an open source license, including it in the official Nematus repository.\footnote{\url{https://github.com/EdinburghNLP/nematus}}
The configuration files needed to replicate our experiments are available in a separate repository.\footnote{\url{https://github.com/Avmb/deep-nmt-architectures}}

\subsection{Layer Normalization}

\begin{table*}
\footnotesize
\centering
\begin{tabular}{l|c|ccc|c|c|c}
encoder & CE & \multicolumn{3}{c|}{\sc Bleu} & parameters (M) & training speed & early stop\\
& & 2014 & 2015 & 2016 & & (words/s) & ($10^4$ minibatches)\\
\hline
baseline & 49.98 & 21.2 & 23.8 & 28.4 & \textbf{\phantom{0}98.0} & \textbf{3350} & 44 \\ 
+layer normalization & \textbf{47.53}  & \textbf{21.9} & \textbf{24.7} & \textbf{29.3} & \phantom{0}98.1 &  2900 & \textbf{29} \\ 
\hline
alternating (depth 4) & 49.25 & 21.8 & 24.6 & 28.9 & \textbf{135.8} & \textbf{2150} & 46 \\ 
+layer normalization & \textbf{46.29} & \textbf{22.6} & \textbf{25.2} & \textbf{30.5} & 135.9 & 1600 & \textbf{29} \\ 
\end{tabular}
\caption{Layer normalization results. English$\to$German WMT17 data.}
\label{results-ln}
\end{table*}

Our first experiment is concerned with layer normalization.
We are interested to see how essential layer normalization is for our deep architectures, and compare the effect of layer normalization on a baseline
system, and a system with an alternating encoder with stacked depth 4.
Results are shown in Table~\ref{results-ln}.
We find that layer normalization is similarly effective for both the shallow baseline model and the deep encoder, yielding an average improvement of 0.8--1 {\sc Bleu}, and reducing training time substantially.
Therefore we use it for all the subsequent experiments.

\subsection{Deep Encoders}

\begin{table*}
\footnotesize
\centering
\begin{tabular}{l|ccc|c|ccc|c|c}
encoder & \multicolumn{3}{c|}{depth} & CE & \multicolumn{3}{c|}{\sc Bleu} & parameters (M) & training speed\\
& s. bidir. & s. forw. & trans. & & 2014 & 2015 & 2016 && (words/s)\\
\hline
shallow & 1 & - & 1 & 47.53  & 21.9 & 24.7 & 29.3 & \phantom{0} 98.1 &  2900 \\ 
\hline
alternating & 4 & - & 1 & \textbf{46.29} & 22.6 & 25.2 & \textbf{30.5} & 135.9 & 1600 \\ 
biunidirectional & 1 & 3 & 1 & 46.79 & 22.4 & \textbf{25.4} & 30.0 & 173.7 & 1500 \\ 
deep transition & 1 & - & 4 & 46.54 & \textbf{22.9} & \textbf{25.4} & 30.2 & \textbf{117.0} & \textbf{1900} \\ 
\end{tabular}
\caption{Deep encoder results. English$\to$German WMT17 data.
Parameters and speed are highlighted for the deep recurrent models.}
\label{results-enc}
\end{table*}

In Table~\ref{results-enc} we report experimental results for different architectures of deep encoders, while the decoder is kept shallow.

We find that all the deep encoders perform substantially better than baseline (+0.5--+1.2 {\sc Bleu}), with no consistent quality differences between each other.
In terms of number of parameters and training speed, the deep transition encoder performs best, followed by the alternating stacked encoder and finally the biunidirectional encoder (note that we trained on a single GPU, the biunidirectional encoder may be comparatively faster on multiple GPUs due to its higher model parallelism).

\subsection{Deep Decoders}

\begin{table*}
\footnotesize
\centering
\begin{tabular}{ll|ccc|c|ccc|ccc|c|c}
decoder & high RNN & \multicolumn{2}{c}{decoder RNN depth} & output & CE & \multicolumn{3}{c|}{\sc Bleu} & params. & training speed\\
& stacked & trans. & type & depth &&2014 & 2015 & 2016 & (M) & (words/s)\\
\hline
shallow & - & 1 & 1 & 1 & 47.53  & 21.9 & 24.7 & 29.3 & \phantom{0} 98.1 &  2900 \\ 
\hline
stacked & GRU & 4 & 1 & 1 & 46.73 & 21.8 & 24.6 & 29.5 & \textbf{117.0} & \textbf{2250} \\ 
stacked & rGRU & 4 & 1 & 1 & 46.72 & 22.1 & 25.0 & 29.4 & 135.9 & 2150  \\ 
stacked & cGRU & 4 & 1 & 1 & \textbf{44.76} & \textbf{22.8} & \textbf{25.5} & 29.6 & 164.3 & 1300 \\ 
stacked & crGRU & 4 & 1 & 1 & 45.88 & 22.5 & 24.7 & 29.7 & 145.4 & 1750 \\ 
deep transition & - & 1 & 8 & 1 & 45.98 & 22.4 & 24.9 & \textbf{30.0} & \textbf{117.0} & 2200 \\ 
\hline
deep output & - & 1 & 1 & 4 & 47.21 & 21.5 & 24.2 & 28.7 & \phantom{0}98.9 & 2850 \\ 
\end{tabular}
\caption{Deep decoder results. English$\to$German WMT17 data.
Parameters and speed are highlighted for the deep recurrent models.}
\label{results-dec}
\end{table*}

Table~\ref{results-dec} shows results for different decoder architectures, while the encoder is shallow.
We find that the deep decoders all improve the cross-entropy, but the {\sc Bleu} results are more varied: deep output\footnote{deep feed-forward output with shallow RNNs in both the encoder and decoder} decreases {\sc Bleu} scores (but note that the baseline has already some depth), stacked GRU performs similarly to the baseline (-0.1--+0.2 {\sc Bleu}) and stacked rGRU possibly slightly better (+0.1--+0.2 {\sc Bleu}).

Other deep RNN decoders achieve higher gains.
The best results (+0.6 {\sc Bleu} on average) are achieved by the stacked conditional GRU with independent multi-step attention (cGRU).
This decoder, however, is the slowest one and has the most parameters.

The deep transition decoder performs well (+0.5 {\sc Bleu} on average) in terms of quality and is the fastest and smallest of all the deep decoders that have shown quality improvements.

The stacked conditional GRU with reused attention (crGRU) achieves smaller improvements (+0.3 {\sc Bleu} on average) and has speed and size intermediate between the deep transition and stacked cGRU decoders.

\subsection{Deep Encoders and Decoders}

\begin{table*}
\scriptsize
\centering
\begin{tabular}{lll|ccc|cc|c|ccc|c|c}
encoder & decoder & decoder high & \multicolumn{3}{c|}{encoder depth} & \multicolumn{2}{c|}{decoder depth} & CE & \multicolumn{3}{c|}{\sc Bleu} & params. & training speed\\
& & RNN type & \multicolumn{3}{c|}{bidir.  forw.  trans.} & stacked & trans. & &2014 & 2015 & 2016 & (M) & (words/s)\\
\hline
shallow & shallow & - & 1 & - & 1 & 1 & 1 & 47.53  & 21.9 & 24.7 & 29.3 & \phantom{0} 98.1 &  2900 \\ 
deep tran. & shallow & - & 1 & - & 4 & 1 & 1 & 46.54 & 22.9 & 25.4 & 30.2 & 117.0 & 1900 \\ 
\hline

\multicolumn{3}{c|}{\citep{zhou2016deep} (ours)} & & & & & & & & & & & \\
alternating & stacked & GRU & 4 & - & 1 & 4 & 1 & 45.89 & 22.9 & 25.3 & 30.1 & 154.9 & 1480 \\  
\hline

\multicolumn{3}{c|}{\citep{wu2016google} (ours)} & & & & & & & & & & & \\
biunidir. & stacked & rGRU & 1 & 3 & 1 & 4 & 1 & 46.15 & 22.4 & 24.7 & 29.6 & 211.5 & 1280 \\ 
\hline

alternating & stacked & rGRU & 4 & - & 1 & 4 & 1 & 46.00 & 23.0 & \textbf{25.7} & 30.5 & 173.7 & 1400 \\ 
alternating & stacked & cGRU & 4 & - & 1 & 4 & 1 & 44.32 & 22.9 & \textbf{25.7} & 29.8 & 202.1 & \phantom{0}970 \\ 
deep tran. & deep tran. & - & 1 & - & 4 & 1 & 8 & 45.52 & 22.7 & \textbf{25.7} & 30.1 & \textbf{136.0} & \textbf{1570} \\ 
BiDeep altern. & BiDeep & rGRU & 2 & - & 2 & 2 & 4/2 & \textbf{43.52} & \textbf{23.1} & 25.5 & \textbf{30.6} & 145.4 & 1480 \\ 
\hline
BiDeep altern. & BiDeep & rGRU & 4 & - & 2 & 4 & 4/2 & \textbf{43.26} & \textbf{23.4} & \textbf{26.0} & \textbf{31.0} & \textbf{214.7} & \textbf{\phantom{0}980} \\ 
alternating & stacked & rGRU & 8 & - & 1 & 8 & 1 & 44.32 & 22.9 & 25.5 & 30.5 & 274.6 & \phantom{0}880 \\ 
\end{tabular}
\caption{Deep encoder--decoder results. English$\to$German WMT17 data.
Transition depth 4/2 means 4 in the base RNN of the stack and 2 in the higher RNNs.
The last two models are large and their results are highlighted separately.}
\label{results-encdec}
\end{table*}

Table~\ref{results-encdec} shows results for models where both the encoder and the decoder are deep, in addition to the results of the best deep encoder (the deep transition encoder) + shallow decoder reported here for ease of comparison.

Compared to deep transition encoder alone, we generally see improvements in cross-entropy, but not in {\sc Bleu}.
We evaluate architectures similar to \citet{zhou2016deep} (alternating encoder + stacked GRU decoder) and \citep{wu2016google} (biunidirectional encoder + stacked rGRU decoder), though they are not straight replications since we used GRU cells rather than LSTMs and the implementation details are different.
We find that the former architecture performs better in terms of {\sc Bleu} scores, model size and training speed.

The other variants of alternating encoder + stacked or deep transition decoder perform similarly to alternating encoder + stacked rGRU decoder, but do not improve {\sc Bleu} scores over the best deep encoder with shallow decoder.
Applying the BiDeep architecture while keeping the total depth the same yields small improvements over the best deep encoder (+0.2 {\sc Bleu} on average), while the improvement in cross-entropy is stronger.
We conjecture that deep decoders may be better at handling subtle target-side linguistic phenomena that are not well captured by the 4-gram precision-based {\sc Bleu} evaluation.

Finally, we evaluate a subset of architectures with a combined depth that is 8 times that of the baseline.
Among the large models, the BiDeep model yields substantial improvements (average +0.6 {\sc Bleu} over the best deep encoder, +1.5 {\sc Bleu} over the shallow baseline), in addition to cross-entropy improvements.
The stacked-only model, on the other hand, performs similarly to the smaller models, despite having even more parameters than the BiDeep model.
This shows that it is useful to combine deep transitions with stacking, as they provide two orthogonal kinds of depth that are both beneficial for neural machine translation.

\subsection{Error Analysis}

One theoretical difference between a stacked RNN and a deep transition RNN is that the distance in the computation graph between timesteps is increased for deep transition RNNs.
While this allows for arguably more expressive computations to be represented, in principle it could reduce the ability to remember information over long distances, since each layer may lose information during forward computation or backpropagation.
This may not be a significant issue in the encoder, as the attention mechanism provides short paths from any source word state to the decoder, but the decoder contains no such shortcuts between its states, therefore it might be possible that this negatively affects its ability to model long-distance relationships in the target text, such as subject--verb agreement.

Here, we seek to answer this question by testing our models on Lingeval97 \citep{sennrich:2017:EACLshort}, a test set which provides contrastive translation pairs for different types of errors.
For the example of subject-verb agreement, contrastive translations are created from a reference translation by changing the grammatical number of the verb,
and we can measure how often the NMT model prefers the correct reference over the contrastive variant.

\begin{figure}
\centering
\begin{tikzpicture}[scale=0.7]
\pgfplotsset{major grid style={style=dotted,color=black!20}}

\pgfplotsset{
    overwrite last x tick label/.style={
        every x tick label/.append style={alias=lasttick},
        extra description/.append code={
            \fill [white] (lasttick.north west) ++(0pt,-\pgflinewidth) rectangle (lasttick.south east);
            \node [anchor=base] at (lasttick.base) {#1};}
    },
    overwrite last x tick label/.default={$\ge$ 16}
}

\begin{axis}[xlabel=distance,
    xmin=0,
    xmax=16,
    ymin=0.8,
    ymax=1,
    ylabel=accuracy,
    xtick={0,4,8,12,16},
        overwrite last x tick label,
    legend pos = south west,
    x=0.5cm,
    legend style={
        /tikz/nodes={anchor=west}
        },
    mark size = 0.1,
    ]

    \addplot +[black, no markers, raw gnuplot, dashed, line width=0.2ex, id=shallow] gnuplot {plot 'plots/exp17.plot' };
    \addplot +[orange, no markers, raw gnuplot, line width=0.15ex, id=stack] gnuplot {plot 'plots/exp33.plot' };
    \addplot +[red, no markers, raw gnuplot, dotted, line cap=round, line width=0.2ex, id=drt] gnuplot {plot 'plots/exp32.plot' };
    \addplot +[blue, no markers, raw gnuplot, dash pattern=on 4pt off 1pt on 4pt off 4pt, line cap=round, line width=0.15ex, id=bideep] gnuplot {plot 'plots/exp35.plot' };

    \addlegendentry{shallow GRU}
    \addlegendentry{stacked GRU}
    \addlegendentry{deep transition GRU}
    \addlegendentry{BiDeep GRU}

\end{axis}
\end{tikzpicture}
\caption{Subject-verb agreement accuracy as a function of distance between subject and verb.}
\label{agreement-by-distance}
\end{figure}
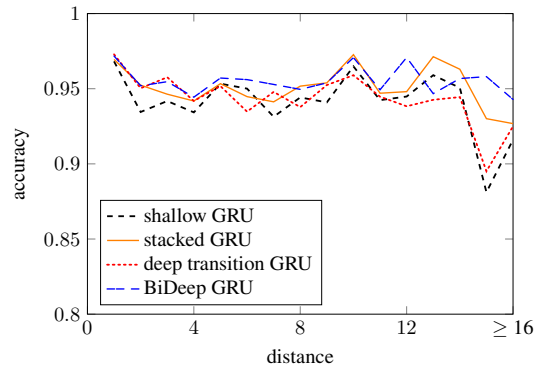

In Figure~\ref{agreement-by-distance}, we show accuracy as a function of the distance between subject and verb.
We find that information is successfully passed over long distances by the deep recurrent transition network.
Even for decisions that require information to be carried over 16 or more words, or at least 128 GRU transitions\footnote{some decisions may not require the information to be passed on the target side because the decisions may be possible based on source-side information.}, the deep recurrent transition network achieves an accuracy of over 92.5\% ($N=560$), higher than the shallow decoder (91.6\%), and similar to the stacked GRU (92.7\%).
The highest accuracy (94.3\%) is achieved by the BiDeep network.

\section{Conclusions}
\label{sec:conclusions}

In this work we presented and evaluated multiple architectures to increase the model depth of neural machine translation systems.

We showed that \textit{alternating stacked} encoders \citep{zhou2016deep} outperform \textit{biunidirectional stacked} encoders \cite{wu2016google}, both in accuracy and (single-GPU) speed.
We showed that \textit{deep transition} architectures, which we first applied to NMT, perform comparably to the stacked ones in terms of accuracy ({\sc Bleu}, cross-entropy and long-distance syntactic agreement), and better in terms of speed and number of parameters.

We found that depth improves {\sc Bleu} scores especially in the encoder.
Decoder depth, however, still improves cross-entropy if not strongly {\sc Bleu} scores.

The best results are obtained by our BiDeep architecture which combines both stacked depth and transition depth in both the (alternating) encoder and the decoder, yielding better accuracy for the same number of parameters than systems with only one kind of depth.

We recommend to use combined architectures when maximum accuracy is the goal, or use deep transition architectures when speed or model size are a concern, as deep transition performs very positively in the quality/speed and quality/size trade-off.

While this paper only reports results for one translation direction, the effectiveness of the presented architectures across different data conditions and language pairs was confirmed in follow-up work.
For the shared news translation task of this year's Conference on Machine Translation (WMT17), we built deep models for 12 translation directions, using a deep transition architecture or a stacked architecture (alternating encoder and rGRU decoder), and observe improvements for the majority of translation directions \cite{uedin-nmt:2017}.

\section*{Acknowledgments}
The  research  presented  in  this  publication  was conducted  in  cooperation  with  Samsung  Electronics  Polska  sp.  z  o.o.  -  Samsung  R\&D  Institute  Poland.
\lettrine[image=true, lines=2, findent=1ex, nindent=0ex, loversize=.15]{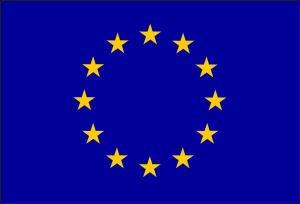}This  project  received  funding from the European Union’s Horizon 2020 research and innovation programme under grant agreements 645452 (QT21), 644402 (HimL) and 688139 (SUMMA).

\bibliography{deep}
\bibliographystyle{emnlp_natbib}

\end{document}